# Good Counterfactuals and Where to Find Them:
# A Case-Based Technique for Generating Counterfactuals
# for Explainable AI (XAI)

Mark T. Keane[1,2,3] & Barry Smyth[1,2]

[1] School of Computer Science, University College Dublin, Dublin, Ireland
[2] Insight Centre for Data Analytics, University College Dublin, Dublin, Ireland
[3] VistaMilk SFI Research Centre, University College Dublin, Dublin, Ireland
{mark.keane, barry.smyth}@ucd.ie

**Abstract.** Recently, a groundswell of research has identified the use of counterfactual explanations as a potentially significant solution to the Explainable AI (XAI) problem. It is argued that (i) *technically*, these counterfactual cases can be generated by permuting problem-features until a class-change is found, (ii) *psychologically*, they are much more causally informative than factual explanations, (iii) *legally*, they are GDPR-compliant. However, there are issues around the finding of "good" counterfactuals using current techniques (e.g. *sparsity* and *plausibility*). We show that many commonly-used datasets appear to have few "good" counterfactuals for explanation purposes. We propose a new case-based approach for generating counterfactuals, using novel ideas about the *counterfactual potential* and *explanatory coverage* of a case-base. The new technique reuses patterns of good counterfactuals, present in a case-base, to generate analogous counterfactuals that can explain new problems and their solutions. Several experiments show how this technique can improve the counterfactual potential and explanatory coverage of case-bases, that were previously found wanting.

**Keywords:** CBR, Explanation, XAI, Counterfactuals, Contrastive

## 1 Introduction

In recent years, there has been a tsunami of papers on Explainable AI (XAI) reflecting concerns that recent advances in machine learning may be limited by a lack of transparency (see e.g., [1, 2]) or by government regulation (e.g., GDPR in the EU, see [3, 4]; for reviews [5, 6, 7, 8]). Historically, Case-Based Reasoning (CBR) has always given a central role to explanation, as predictions can readily be explained by cases, akin to human reasoning from precedent/example [9, 10, 11]; see [11-15]). Indeed, Keane & Kenny's [16, 17] *twin systems approach*, explicitly maps opaque deep-learning systems into CBR systems to find *post-hoc* explanatory cases for their predictions. Typically, CBR uses "factual cases" (i.e., nearest *like* neighbors that provide evidence for why a prediction was made, see [16, 17]). But, recently, there has been explosion of interest in a different class of explanatory cases, so-called *counterfactual cases* (i.e., nearest *unlike* neighbors that explain how a prediction might be changed). For example, a loan application system might explain its decision to refuse a loan by presenting a factual case: "*you were refused because a previous customer had the same salary as you and they were refused a loan for this amount*". In contrast, same loan system might, arguably, provide a better explanation by presenting a counterfactual case; effectively saying "*if you asked for a slightly lower amount you would have been granted the loan*". Researchers championing the use of counterfactual explanations, argue that they provide better solutions to the XAI problem [7, 18, 19, 20, 21] (see section 2).

In this paper, we consider the feasibility of counterfactual explanations from a CBR perspective. Though any CBR system could explain its predictions directly using counterfactual cases, in the present work, we are assuming a twin-system context [16, 17]; where there is an opaque machine-learning model (e.g., deep learning system) that is generating predictions that have to be explained by finding case-based explanations from a twinned CBR system[1]. In this context, we assess how many "good" counterfactuals are available in a given case-base (i.e., ones that are easily comprehended by people). So, we systematically map the topology of "good" counterfactuals in different case-bases, what we call their *counterfactual potential* (see section 2). Initially, we perform an analysis of 20 frequently used case-bases from the CBR literature (see section 3). To our surprise, to presage our results, we find that in most case-bases "good" counterfactuals are quite rare. This leads us to the novel notion of *explanatory coverage* by analogy to *predictive coverage* [22, 23, 24], from which we develop and evaluate a new case-based technique for counterfactual generation in XAI (sections 4-5).

## 2    Counterfactual Explanation: Promise, Problems & Prospects

Intuitively, counterfactual explanations seem to provide better explanations than factual ones; in CBR-ese, nearest-unlike-neighbor (NUN) explanations are better than nearest-like-neighbor (NLN) explanations[2]. Imagine you are at a party and you want to know whether you are okay to drive home; you have the latest DeepDrink app that can predict whether you are under or over the alcohol limit for driving. DeepDrink knows your physical details and, after you tell it how many drinks you have taken, your recent food consumption and when you started drinking, it predicts that you are *over the limit* explaining its prediction using a *factual case,* saying that a person with a similar profile to you was also over the limit when they were breathalysed (see Table 1). This explanation is reasonable but perhaps not as good as one using a *counterfactual case*; which would tell you that someone with your profile who drank a similar amount over a longer period, ended up being under the limit (see "good" counterfactual in Table 1). The counterfactual directly tells you more about the causal dependencies in the domain and, importantly, provides you with "actionable" information (i.e., that if you stopped drinking for 30 minutes you could be under the limit). Technically, counterfactuals can tell

---

[1]   Thus, this context assumes an existing (albeit opaque) model to which cases can be presented to find predictions/labels. Access to such a model is assumed by all counterfactual-generation techniques, though there is some discussion around whether the training data would also always be accessible (obviously, we assume that training-data/case-base is available).

[2]   We have found few CBR papers that seriously consider counterfactual cases (aka NUNs) for explanation. Obviously, NUNs have been studied in CBR, but typically not for explanation (e.g., [25, 26]).  [27-29] did consider NUNs in an explanation context, but more as confidence indicators with respect to decision boundaries.

you about the feature differences that affect the decision boundary around a prediction. Accordingly, [20] define counterfactual explanations as statements taking the form:

*Score y was returned because variables V had values (v1, v2,.. ). If V had values (v1', v2' ... ), and all others remain constant, score y' would have been returned.*

where, in our example, score *y* would be the class "over the limit" and *y'* the class "under the limit". Recently, researchers championing counterfactual cases for XAI have argued that psychologically, technically and legally they provide better explanations than other techniques for XAI [7, 19, 20, 30, 31].

Table 1. A Test Case paired with a "Good" and a "Bad" Counterfactual from the Blood Alcohol Content (BAC) case-base with the feature-differences between them (shown in bold italics)

| Features | Test Case | "Good" Counterfactual | "Bad" Counterfactual |
|---|---|---|---|
| *Weight* | 80 kg | 80 kg | 80 kg |
| *Duration* | 1 hr | *1.5 hrs* | *3 hrs* |
| *Gender* | Male | Male | *Female* |
| *Meal* | Empty | Empty | *Full* |
| *Units* | 6 | 6 | *6.5* |
| *Bac Level* | Over | Under | Under |

## 2.1 Counterfactual Promise

Many have argued that counterfactual thinking has a promising role to play in explanation from philosophical, psychological, computational and legal perspectives. Philosophers of science have argued that true causal explanation only emerges from contrastive propositions, using counterfactuals (see e.g., [32, 33]). Psychologists have also shown that counterfactuals play a key role in human cognition and emotion, eliciting spontaneous causal thinking about what might have been the case (e.g. [18, 19, 34]). Byrne [19] has explicitly related this literature to the XAI problem, laying out the different ways in which counterfactuals could be used (see also [7, 35]). For example, as counterfactuals engender more active causal thinking in people, they are more likely to facilitate "human in the loop" decision making [19]. Recently, Dodge et al. [36] assessed explanations of biased classifiers using four different explanation styles and found counterfactual explanations to be the most effective. In AI, Pearl [31] has proposed an influential structural Bayesian approach to counterfactuals that has been used to test the fairness of AI systems, but it has been less used in explanation generation (e.g., see [37, 38, 39]). In the XAI literature, the use of counterfactuals has emerged as an active counterpoint to popular post-hoc perturbation approaches (e.g., LIME, [40, 41]; and also [42]), with many researchers arguing that counterfactuals provide more robust and informative post-hoc explanations [21, 30, 43-45]; these "counterfactualists" have also argued that counterfactual explanations are GDPR compliant [4, 20, 44].

## 2.2 Counterfactual Problems

However, the promise of counterfactuals for XAI comes with a number of problems; the three main ones being prolixity, sparcity and plausibility.

**Prolixity.** Currently, most XAI systems generate counterfactuals using random perturbation and search, making them very *prolix* [4, 20]; that is, many different counterfactuals may be produced for a given test case from which a "good" one must be selected (e.g., in the loan system, one could be shown counterfactuals for every 10$ incremental change in one's salary, but this would not be very helpful). Stated simply, this prolixity is reduced by using methods that find the minimal changes to the features of the test case that flip the prediction (i.e., the nearest unlike neighbor). So, [20] propose the following loss function, $L$:

$$L(x, x', y', \lambda) = \lambda(f(x') - y')^2 + d(x, x') \qquad (1)$$

$$arg \min_{x'} \max_{\lambda} L(x, x', y', \lambda) \qquad (2)$$

$$d(x, x') = \sum_{j=1}^{p} \frac{|x_j - x'_j|}{MAD_j}, \qquad (3)$$

$$MAD_j = median_{i \in \{1, \dots, n\}}(|x_{i,j} - median_{l \in \{1, \dots, n\}}(x_{l,j})|) \qquad (4)$$

where $x$ is the vector for the test case and $x'$ is the counterfactual vector, with $y'$ being the desired (flipped) prediction from $f(..)$ the trained model, where $\lambda$ acts as the balancing weight. In formula (2), $\lambda$ balances the closeness of the counterfactual to the test case against making the smallest possible changes to the test case while delivering a prediction change, using the distance metric in (3) and (4) which is Manhattan Distance ($L1$) using the median absolute deviation (MAD) of each feature. While different researchers optimize this function in different ways [20, 44, 46] tests show that this technique finds minimally-mutated counterfactuals to the test case, solving the prolixity problem (see also [44] work on *diversity* between generated counterfactuals).

**Sparcity.** These methods also profess to solve the *sparcity* problem. All commentators argue that good explanatory counterfactuals need to be *sparse*; that is, they need to modify the *fewest* features in the selected counterfactual. For example, Table 1 shows, for the blood alcohol domain, two different counterfactuals, one with a 1-feature change and another with a 4-feature change, where the sparsity of the former makes it better than the latter. So, [4] argue their MAD distance metric delivers sparse counterfactuals (see above), although tests [46] show that many of these counterfactuals may still involve relatively high numbers of feature-differences (e.g., >4). Importantly, the argument for sparcity is a psychological one that has not been specifically tested in the XAI literature. Typically, AI researchers propose sparcity is important because of human working memory limits [47, 48], but we believe that people prefer sparse counterfactuals because of constraints on human category learning. For example, [49] have shown that, in concept learning, when people are trying to learn categories for unfamiliar items they prefer single-feature changes between to-be-learned items over multiple-feature changes, because it makes the learning task easier (unless there is additional domain knowledge showing dependencies between features). Based on this evidence, we operationalize the *sparcity* of "good" counterfactuals (as items with 1 or 2 feature differences) versus "bad" counterfactuals (those with >2 feature changes). This definition allows us to develop the novel idea of the *counterfactual potential* of case-bases, based on quantifying the "good" counterfactuals they contain (see sections 4-5).

**Plausibility.** The final problem is that of *plausibility*; that is, the counterfactuals generated may not be valid data-points in the domain or they may suggest feature-changes that are difficult-to-impossible. For example, classic cases of such counterfactuals in the loans domain, are explanations that propose increasing one's salary by an implausible amount (i.e., *if you earned $1M, you would get the loan*) or quite radical proposals (i.e., *if you changed your gender, you would get the loan*). Plausibility is the least solved of the three problems facing counterfactuals; for instance, many researchers propose to "lock" features (e.g., to not allow *gender* change) or to allow users to provide inputs on feature weights [44] (e.g., using sliders at the interface on, say, salary boundaries). However, attempts to find an automated solution to the plausibility problem are thin on the ground[3]. Here, we propose that, rather than generating counterfactuals by "blind" random perturbation, an XAI system should use the training-data/case-base to find suitable counterfactuals; as these counterfactuals are "real experiences" from the problem domain, they have an inherent plausibility. However, this leads us to another question: namely, how many good counterfactuals are "naturally" available in case-bases or, more simply, what is their *counterfactual potential*.

### 2.3    CBR's Prospects for Counterfactuals

Most techniques for generating counterfactuals for XAI perform random perturbations of a query-case followed by search to find a minimally-different case that is close to the decision boundary (i.e., a NUN). These perturbation techniques can encounter problems, notably in meeting *sparcity and plausibility,* which may benefit from a case-based approach. Just as CBR has successfully explained predictions using factual cases [29, 12], perhaps it can also deliver counterfactual cases that are *sparse* (selected using a suitable similarity metric) and *plausible* (because they are based on previously encountered training data). However, if CBR is to be used, we need to establish whether case-bases actually contain good counterfactuals, whether they have high counterfactual potential.   In what follows we define a *good counterfactual* to be a NUN that differs from the query case by no more than 2 features. This potential can be determined by computing the feature-differences in all pairwise comparisons of cases in the case-base. If these comparisons show there are many "good" counterfactuals then the potential is high, if not then it is low. So, in our first experiment, we computed the counterfactual potential of 20 classic datasets, used in the CBR literature, from the UCI repository [51]. From this analysis we develop the idea of *explanatory coverage* before proposing a novel case-based technique for counterfactual generation (section 4). Finally, in section 5, we report a set of experiments on five representative datasets to show how the technique can improve counterfactual potential.

## 3    Experiment 1: Plotting Counterfactual Potential

In this experiment, we computed the counterfactual potential of 20 classic datasets from the UCI repository [51], ones that have been commonly used in 10s if not 100s of CBR papers. This analysis was done by computing the number of feature differences between all pairwise comparisons of cases in the case-base, noting the proportion of "good" counterfactuals found (i.e., ≤2 feature difference counterfactuals). This analysis provides us with an upper/lower bound on the potential of a case-base to deliver good counterfactuals. Obviously, in any specific CBR system, one might be able to adjust weights, feature-matches or $k$-values to find such counterfactuals, but such fine-tuning will not improve matters hugely if good counterfactual-cases are just not there.

---

[3]    One notable exception is [50] who try to "justify" generated counterfactuals with respect to nearest neightbours from the training set.

**Table 2.** Percent Counterfactuals for Feature-Differences in 20 UCI Datasets (Expt.1)

| DataSets | N of cases | Feat. No. | Class No | N of Pairs | 1-diff | 2-diff | 3-diff | 4-diff | >5-diff |
|---|---|---|---|---|---|---|---|---|---|
| Abalone | 4177 | 10 | 8 | 15.6M | 0% | 0% | 0% | 0% | 99.9% |
| Auto MPG | 398 | 8 | 5 | 52.3k | 0% | 0% | 0% | 0.4% | 99.6% |
| BAC | 9291 | 7 | 2 | 19M | 0% | 1.5% | 23% | 3% | 72% |
| Bupa liver | 345 | 6 | 2 | 29k | 0% | 0% | 0.1% | 3.1% | 96.8% |
| Credit | 653 | 15 | 2 | 105.7k | 0% | 0% | 0% | 0% | 99.9% |
| Cleveland heart | 303 | 13 | 5 | 32.9k | 0% | 0% | 0% | 0.1% | 99.9% |
| Ecoli | 336 | 7 | 7 | 41k | 0% | 0% | 0% | 0.2% | 99.8% |
| Glass | 214 | 9 | 7 | 21.9k | 0% | 0% | 0% | 0% | 99.9% |
| German credit | 914 | 20 | 2 | 177k | 0% | 0% | 0% | 0% | 99.9% |
| Horse colic | 300 | 22 | 2 | 20.8k | 0% | 0% | 0% | 0% | 99.9% |
| Indian liver | 583 | 10 | 2 | 69.5k | 0% | 0% | 0% | 0% | 99.9% |
| Ionosphere | 351 | 34 | 2 | 28.3k | 0% | 0% | 0% | 0% | 100% |
| Iris | 150 | 4 | 3 | 7.5k | 0% | 0.3% | 8.8% | 91% | - |
| Sonar | 208 | 60 | 2 | 10.8k | 0% | 0% | 0% | 0% | 100% |
| Soybean (large) | 307 | 26 | 19 | 43k | 0% | 0% | 0.2% | 0.6% | 99.2% |
| Thyroid | 2800 | 27 | 3 | 355.8k | 0% | 0% | 0% | 0% | 99.9% |
| Votes | 435 | 17 | 2 | 44.8k | 0% | 0.3% | 0.9% | 1.9% | 88.8% |
| Wine-Italian | 178 | 13 | 3 | 10.4k | 0% | 0% | 0% | 0% | 100% |
| Wisconsin breast | 699 | 9 | 2 | 110k | 0% | 0% | 0% | 0.4% | 99.5% |
| Yeast | 1484 | 8 | 10 | 855.3k | 0% | 0% | 0.3% | 4.8% | 94.9% |

### 3.1   Method: Data Sets & Procedure

Twenty UCI datasets were used in the experiment, selected on the basis of their common usage in CBR. We compared all pairings of test cases (one side of a decision boundary) to training cases (on the other side of a decision boundary) calculating the the number of feature differences found in each.

### 3.2   Results & Discussion

Table 2 shows the counterfactual potential of these UCI datasets, as the percentage of counterfactuals with 1 upto >5 feature-differences. These results show that "good" counterfactuals are rare[4]; in nearly every dataset, the 1-diff and 2-diff counterfactual categories account for < 1% of the total collection of counterfactuals. Most counterfactuals involve >5 feature-differences, showing poor sparsity.

It should be noted that in the above, we determine feature differences using a exact-match approach. Such an approach is inherently conservative in the case of real-valued features. In practice, this can be addressed by introducing a degree of matching tolerance so that, for example, we treat two feature values as equivalent if they are within 1% of each other. While this improves counterfactual availability, we still found that the fraction of good counterfactuals (≤2 feature differences) typically remains very low.

On the face of it, this suggests that perhaps a case-based approach to counterfactual generation is a bad idea. If most case-bases do not deliver up many good counterfactuals then case-based techniques may be bound to fail? However, as we shall see in the following sections, there are additional steps that can be applied to meet these challenges.

## 4 A Case-Based Technique for Good Counterfactuals

The above analysis of case-bases suggests that, ironically, CBR seems to have little to offer with respect to the use of counterfactuals in XAI. For most case-bases good counterfactuals are rare and/or are hard to find; that few query/problem-cases will have an associated good counterfactual-case. Perhaps this explains why the dominant technique in the literature uses perturbation, where synthetic counterfactuals are generated "blindly" from problem-cases and labelled using the available machine-learning model, without reference to other known cases in the training set [21, 30, 43, 44, 45, 46]. In contrast to these approaches, we believe that counterfactuals need to be explicitly grounded in known cases (*aka* the training data), to ensure plausibility. Hence, we have developed a novel case-based technique for counterfactual-XAI which re-uses patterns of good counterfactuals, that already exist in a case-base, to generate analogous counterfactuals that are suitable to explain new target problems and their solutions. When it comes to generating new counterfactuals, these existing good counterfactuals provide 'hints' about what features can and should be adapted and plausible feature values to use for them. This new technique relies on the notion of *explanatory competence* (see 4.1). As we said earlier, the context for the use of this method is a twin-system approach to XAI, where an opaque model (e.g., a deep learning network) is "explained" by twinning it with a more transparent CBR-system to find explanatory cases [16,17]; so, along with all other counterfactual-generation techniques, we assume that an ML model is available to assign labels for any newly-generated case.

---

[4] We extensively tested the Blood Alcohol Content (BAC, [27-29]) case-base, but cannot report it for reasons of space. Using a mechanical model for estimating BAC, we generated many different master-case-bases from which we sampled 50+ specific case-bases, all of which repeatedly showed the same absence of good counterfactuals.

### 4.1 Explanatory Competence

The notion of *predictive competence* or simply *competence* (i.e., an assessment of an ML/CBR system's potential to solve a range of future problems) has proved to be a very useful development for AI systems [24]. For example, in CBR, predictive competence has been used to assess the overall problem-solving potential of a system, to help avoid the utility problem as a case-base grows, to maintain case-bases and so on [22, 23]. A parallel notion of *explanatory competence* can also be applied to any case-base.

Just as the fundamental unit of (predictive) competence is a relation of the form *solves(c, c')* to indicate that case/example $c$ can be used to solve some target/query $c'$, so too the basic unit of explanatory competence is *explains(c, c')* indicating that some case $c$ can be used to explain the solution of $c'$; in this work, the explanatory cases ($c$) are the counterfactuals of $c'$. Accordingly, the explanatory competence of a case-base $C$ can be represented by a *coverage set* (Eq. 1) and degree of explanatory competence can be estimated as the size of the coverage set as a fraction of the case-base (Eq. 2).

$$XP\_Coverage\_Set(C) = \{ c' \in C \mid \exists c \in C\text{-}\{c'\} \; \& \; explains(c, c')\} \qquad (1)$$

$$XP\_Coverage(C) = |XP\_Coverage\_Set(C)| \, / \, |C| \qquad (2)$$

### 4.2 Leveraging Counterfactual Cases for Explanation

Although good counterfactuals are rare, in practice we can expect most case-bases to offer some examples where a query/problem case can be associated with a good counterfactual, with or without some matching tolerance as mentioned previously. For example, in the Abalone dataset, even though there are few good counterfactuals, using a similarity tolerance of 0.02 means that just over 20% of cases can be associated with good counterfactuals; for the Liver a tolerance of 0.025 means that 4% of cases can be associated with good counterfactuals. Can pairs of cases and their corresponding good counterfactuals guide the search for novel (good) counterfactuals for new target problems that are otherwise without a good counterfactual?

In what follows we will refer to the pairing of a case and its corresponding good counterfactual as an *explanation case* or *XC*. For any given case-base we can generate a corresponding case-base of these explanation cases for use during counterfactual generation; see Eqs. 3 and 4. By definition explanation cases are symmetric — either of the cases can be viewed as the query or counterfactual, which, in practice, means that each pair of unlike neighbours, which differ by $\leq 2$ features, contributes two XCs to the XC case-base.

$$xc(c, c') \Leftrightarrow class(c) \neq class(c') \; \& \; diffs(c, c') \leq 2 \qquad (3)$$

$$XC(C) = \{(c, c') : c, c' \in C \; \& \; xc(c, c')\} \qquad (4)$$

Each XC is associated with a set of *match-features (m)*, the features that are the same between the query and counterfactual (using a specified tolerance), and a set of *difference-features (d)*, the $\leq 2$ features that differ between the query and counterfactual.

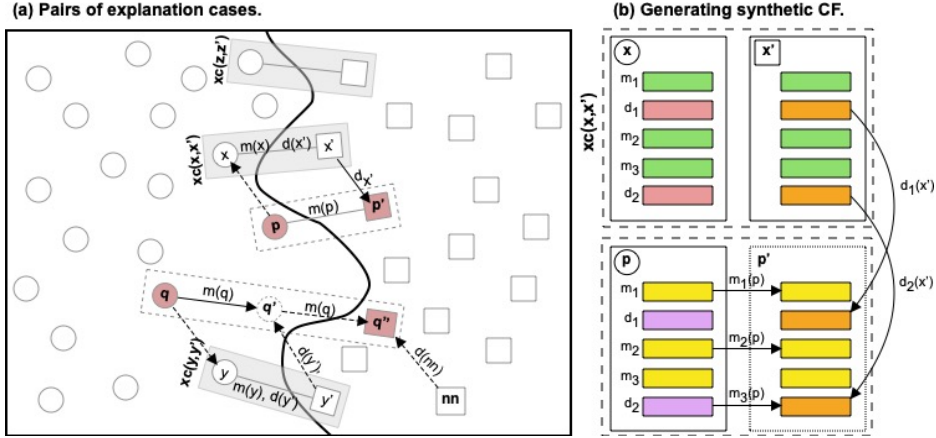

Figure 1. An illustration of (a) a two-class case-base with 3 explanation cases; (b) how a synthetic counterfactual, *(p, p')*, is generated from an existing explanation-case, *xc(x, x')*.

Figure 1(a) shows a two-class case-base of cases (*C*) along with its corresponding *XCs* –– *xc(x, x')*, *xc(y, y')*, and *xc(z, z')* – along with two query cases (*p* and *q*), which have been classified by the underlying ML-model, and which now need to be explained. For our purposes, we assume that there are no existing good counterfactuals for *p* or *q* in *C*, hence the need to generate new good counterfactuals for them.

### 4.3    A Case-Based Approach to Generating Good Counterfactuals

We propose a classical case-based reasoning approach to generating good counterfactuals by *retrieving*, *reusing*, and *revising* a nearby explanation case as follows:

1. First, we identify the XC case whose query is most similar to *p*; this is *xc(x, x')* in Figure 1. Since *xc(x, x')* has a good counterfactual, *x'*, and because the *p* is similar to *x*, then the intuition is that *x'* is a suitable basis for a new counterfactual *p'* to explain *p*. The *difference-features* between *x* and *x'*, which are solely responsible for the class change between *x* and *x'*, should play a critical role in constructing *p'*.

2. For each of the *match-features* in *xc(x,x')*, we copy the *values* of these features in *p* to the new counterfactual *p'*. Similarly, for each of the *difference-features* in *xc(x, x')* we copy their *values* from *x'* into *p'*. In this way, *p'* is a combination of feature values from *p* and *x'*. It differs from *p* in a manner that is similar to the way in which *x'* differs from *x* and, by construction, *p'* is a *candidate good counterfactual* because these differences amount to no more than two features. This transfer of values from *p* and *x'* into *p'* is illustrated in Figure 1(b).

3. For *p'* to be *actually a* good counterfactual, it has to be a different class from *p*, which is not yet guaranteed. We determine the class of *p'* by using the underlying ML-model (from the twin-system)  and if it is different from *p* then *p'* can be used directly as a good counterfactual to explain *p* (see Figure 1(a)).

4. Sometimes, however, the class of the new counterfactual, after retrieval/reuse, is not different from the target query. For example, the new counterfactual $q'$, which is generated for $q$ by reusing $xc(y, y')$ in Figure 1a, has the same class as $q$, because the combination of the match-feature values (from $q$) and difference-features (from $y'$) are not sufficient to change its class from that of $q$.

5. Since $q'$ is not a valid counterfactual, we perform an *adaptation* step to *revise* the values of the difference-features in $q'$ until there is a class change; note, we cannot change the match-features in $q'$ without increasing the number of feature differences with $q$. We can revise the values of the difference-features in $q'$ in various ways, for example, by perturbing them to further increase their distance from $q$. However, we instead iterate over the *ordered nearest neighbours* of $q$ with the same class as $y'$, until there is a class change[5]. The values of the difference features from each *nearest neighbour* leads to a new candidate, $q''$, and adaptation terminates successfully when the class of $q''$ differs from that of $q$; *if* none of the *neighbours* produce a class change, then adaptation fails. In Figure 1(a), when the difference-feature values from the neighbour, *nn*, are used to produce $q''$, the result is a class change, and so $q''$ can be used as a good counterfactual for $q$.

Note that the primary contribution of explanation cases is to identify and distinguish between common combinations of features (match-features and difference-features) that tend to participate in good counterfactuals. Depending on the domain this may reflect some important relationships (causal or otherwise) that exist within the feature-space. In other words, the XCs tell us about which features *should* be changed (or held constant) when generating new counterfactuals in the feature space near a query case.

Another advantage of this approach is that because it reuses *actual feature values* from *real cases,* it should lead to more plausible counterfactuals and, better explanations. This contrasts with perturbation approaches, which rely on arbitrary values for features (and may even produce invalid data-points).

Finally, though our approach may succeed in finding a suitable counterfactual without the need for the adaptation/revision step, it may be desirable to proceed with this step, nonetheless. This is because the adaptation step has the potential to locate a suitable counterfactual that is *closer to the query* than the candidate counterfactual produced by the retrieval step alone and finding counterfactuals that are maximally similar to the query is an important factor when it comes to explanation [20].

## 5 Experiment 2: Evaluating Explanation Competence

We provide a preliminary evaluation of the above approach using five popular ML/CBR datasets to demonstrate how explanatory competence can be improved above and beyond the baseline level of good counterfactuals that naturally occurs in a dataset.

---

5   For multi-class datasets, this adaptation could be modified to iterate over all ordered nearest neighbours with a *different class* to $q$, not just those with the same class as $y'$. This would provide a larger pool of difference-feature values and increase the likelihood of locating a good counterfactual for $q$. We leave this variation, and its evaluation, for future work.

### 5.1    Method: Data & Procedure

Each of the datasets represents a classification task of varying complexity, in terms of the number of classes, features, and training examples. The task of interest, however, is not a classification one but an explanation one. As such we are attempting to generate good counterfactuals in order to *explain* target/query cases and their classes. The key evaluation metrics will be: (a) the fraction of target/query cases than can be associated with good counterfactuals (*explanatory competence*); and (b) the distance from the target/query to the newly generated good counterfactual (*counterfactual distance*).

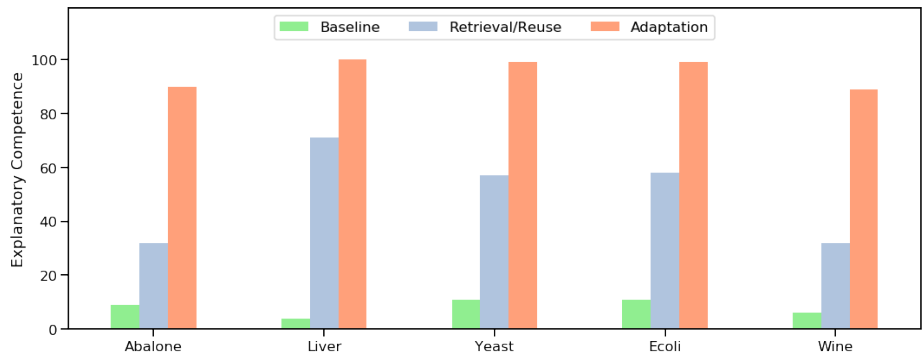

Figure 2. The explanatory competence (XP_Coverage) of five case-bases, showing baseline competence and how competence increases by reusing and adapting explanation cases.

As a baseline for explanatory competence we use the fraction of cases that can be associated with a good counterfactual in each case-base; in each case we use a matching tolerance of 1-2%. As a corresponding baseline for counterfactual distance, we will use the average distance between these cases and their good counterfactuals. A 10-fold cross-validation was used to evaluate the newly generated counterfactuals, selecting 10% of the cases at random to use as queries, and building the XC case-base from the remaining cases. Then, we use the above technique to generate good counterfactuals for the queries, noting the fraction of the queries that can be associated with good counterfactuals, and the corresponding counterfactual distances, after the retrieval/reuse and adaptation steps. Results reported are the averages for the 10 folds for each dataset.

### 5.2    Results & Discussion: Explanatory Competence

The explanatory competence results are presented in Figure 2, showing the explanatory competence (fraction of queries that can be explained) for the dataset (baseline), and for the synthetic counterfactuals generated after the retrieval and adaptation steps of our approach. The results show how explanatory competence can be significantly increased by the case-based-counterfactual technique. For example, on average only about 11% of the cases in these datasets can be associated with good counterfactuals (the average baseline competence when a tolerance is applied) but by retrieving and reusing explanation cases we can reach an average explanatory competence of just over 40%. Implementing the adaptation step further increases the explanatory competence just under 94%, on average. Notably, even datasets with very low baseline explanatory competence benefit from significant improvements in explanatory competence particularly when the adaptation step is used. For example, the 6,400 case Wine dataset (12 features and 7 classes) has a baseline explanatory competence of just 6%, but its 559 XC cases can be used to achieve almost 90% in explanatory competence.

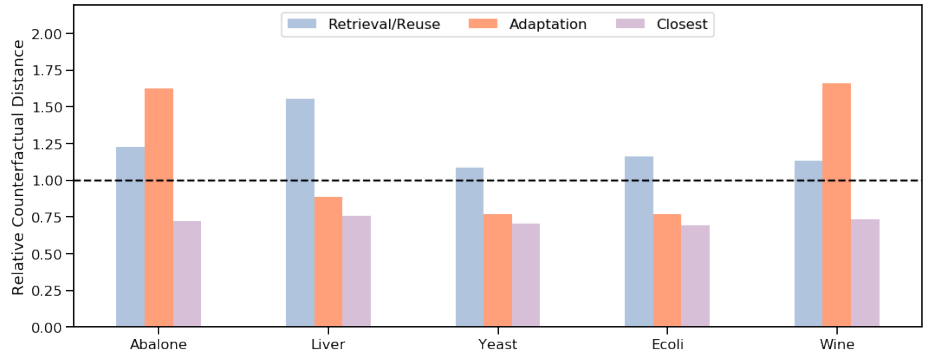

Figure 3. The counterfactual distance of the good counterfactuals produced for five case-bases, relative to the baseline counterfactual distance.

### 5.3 Results & Discussion: Counterfactual Distance

Of course, just because it is possible to generate a counterfactual for a query that has no more than 2 feature differences, does not necessarily mean that the counterfactual will make for an ideal explanation, in practice. To test this would require a succession of live user-trials, beyond the scope of the present work. As an alternative, however, we can use the distance between the query and the generated counterfactual as a proxy for the utility of the explanation, on the grounds that counterfactuals which are closer to a query are more likely to serve as more useful explanations. Since counterfactual distance will vary from dataset to dataset, reflecting the nature of the feature space, in this analysis we present a *relative counterfactual distance* (RCF) measure by dividing the counterfactual distances of the synthetic counterfactuals by the baseline counterfactual distance for the dataset. Thus, if RCF>1, then it indicates that the synthetic counterfactual is farther from the query that the average baseline counterfactual distance.

The results are presented in Figure 3, which include the relative distance of the good counterfactuals produced by the retrieval/reuse and the adaptation steps for each dataset. We also show the relative distance results for an additional condition, *Closest*, which is defined as follows: when both the retrieval/reuse and adaptation steps lead to a good counterfactual, then choose the one with the lower counterfactual distance, otherwise if only one good counterfactual is produced then use its distance.

On average, good counterfactuals produced by the retrieval/reuse step are farther from the test query than the baseline counterfactual distance (RCF $\approx$ 1.2). In most cases the additional distance beyond the baseline is modest with the exception of the Liver dataset, where the retrieval/reuse step produces good counterfactuals that are 55% (RCF $\approx$ 1.55) more distance from the query than the baseline distance. The good counterfactuals produced by the adaptation step are closer to the test queries – the average

RCF $\approx 1.1$, and in 3 out of the 5 datasets the generated counterfactuals are closer than the baseline (RCF $<1$). If we select the closest counterfactual, when both retrieval/reuse and adaptation produce one, then the RCF$<1$ for all of the datasets. This further validates the need for, and quantifies the benefits of, the adaptation step: it provides an opportunity to choose a counterfactual that is significantly closer to the query.

## 6     Conclusions & Future Directions

In the last three years, there has been a significant upsurge in XAI research arguing for the computational, psychological and legal advantages of counterfactuals. Most of this work generates synthetic counterfactuals without reference to the training-data in the domain and, as such, can suffer from *sparsity* and *plausibility* deficits. In short, these methods do not guarantee the production of good counterfactuals and, may indeed, sometimes generate invalid data points. This state of affairs invites a case-based solution to counterfactual generation that leverages the prior experience of the case-base, adapting known counterfactual associations between query-problems and known cases. In this paper, we advance just such a technique and show how it can improve the counterfactual potential of many datasets. In developing this technique, we have (i) clarified the definition of good counterfactuals, (ii) proposed the new idea of explanation competence, (iii) reported significant new evidence for the utility of this novel technique.

The approach, as described, makes some assumptions that might limit its utility beyond the datasets discussed. It assumes the availability of at least some explanation cases, which is typically feasible; even though good counterfactuals are rare they are seldom so rare as exclude minimally viable explanation case-base, at least when a degree of matching tolerance is allowed for when computing feature similarities and differences. The approach also assumes the availability of an underlying ML-model (e.g., in twin system) for the purpose of counterfactual validation, though this is an accepted proposition in all approaches. Finally, though previous psychological work supports our operational definition of good counterfactuals, more user testing is required. Notwithstanding this future research, from the current findings, it is clear that a CBR approach to counterfactual XAI has much to offer.